\documentclass[a4paper]{easychair}
\usepackage{doc}
\usepackage{makeidx}
\usepackage{amssymb}
\usepackage[figuresright]{rotating}
\usepackage{url,lineno}

\def\procediaConference{International Conference on Computational Science 2017 }

\title{\bf Spectral Modes of Network Dynamics Reveal Increased Informational Complexity  Near Criticality  }

\titlerunning{Network Informational Complexity Near Criticality }

\author{
Xerxes D. Arsiwalla\inst{1}  
\and
Pedro A.M. Mediano\inst{2}
\and
   Paul F.M.J. Verschure\inst{1,3}\\
}

\institute{
  Synthetic Perceptive Emotive and Cognitive Systems (SPECS) Lab, Center of Autonomous Systems and Neurorobotics,  Universitat Pompeu Fabra,  Barcelona, Spain.\\
  \email{x.d.arsiwalla@gmail.com}
\and
Department of Computing, Imperial College London. \\
\and
   Instituci\'{o} Catalana de Recerca i Estudis Avan\c{c}ats (ICREA), Barcelona, Spain.\\
}

\authorrunning{ }

\begin{document}

\maketitle

%%%%\keywords{Network dynamics, Complexity measures, Information theory}

\begin{abstract}
What does the informational complexity of dynamical networked systems tell us about intrinsic mechanisms and functions of these complex systems? Recent complexity measures such as integrated information have sought to operationalize this problem taking a whole-versus-parts perspective, wherein one explicitly computes  the amount of information generated by a network as a whole over and above that generated by the sum of its parts during state transitions.  While several numerical schemes for estimating network integrated information exist, it is instructive to pursue an analytic approach that computes integrated information as a function of network weights. Our formulation of integrated information uses a Kullback-Leibler divergence between the multi-variate distribution on the set of network states versus the corresponding factorized distribution over its parts. Implementing stochastic Gaussian dynamics, we perform computations for  several prototypical network topologies. Our findings  show increased informational complexity near criticality, which remains consistent across network topologies.  Spectral decomposition of the system's dynamics reveals how  informational complexity is governed by  eigenmodes of both, the network's covariance and adjacency matrices.   We find that as the dynamics of the system approach  criticality,  high integrated information is exclusively driven by the eigenmode corresponding to the leading eigenvalue of the  covariance matrix, while sub-leading modes get suppressed.  The implication of this result is that it might be favorable for complex dynamical networked systems such as the human brain or  communication systems to operate near  criticality so that efficient information integration might be achieved. 

\end{abstract}

%\setcounter{tocdepth}{2}
%{\small
%\tableofcontents}

%\section{To mention}
%
%Processing in EasyChair - number of pages.
%
%Examples of how EasyChair processes papers. Caveats (replacement of EC
%class, errors).

%\pagestyle{empty}

%------------------------------------------------------------------------------ 

\section{Introduction}
\label{}
Quantifying informational processes of dynamical networked systems has been increasingly useful as  a unique window for probing internal system states and mechanisms that underlie  observed phenomenological behaviors of many complex systems. Mapping structure-function relationships in this way by using information theory has paid off for studying both, complex biological systems such as the brain or engineered systems such as communication networks. A prominent information-theoretic complexity measure that has shown a recent resurgence of interest in the wake of consciousness research is integrated information (often denoted as $\Phi$). It was first introduced in neuroscience as a complexity measure for neural networks, and touted as a  correlate of consciousness  \cite{tononi1994}. Integrated information  $\Phi$  is loosely defined as the quantity of information generated by a network as a whole,  due to its causal dynamical interactions,  that is over and above the information generated independently by the disjoint sum of its parts.  As a complexity measure, $\Phi$  seeks to operationalize the intuition   that complexity arises from simultaneous integration and differentiation of the network's structure and dynamics.  Integration results in distributed coordination among nodes, while differentiation leads to functional specializations,  thus enabling the emergence of the system's collective states.  The interplay between  integration and differentiation thus generates information that is highly diversified yet integrated, creating patterns of high complexity. Following initial proposals \cite{tononi2004}, \cite{tononi2003}, \cite{tononi1994}, several approaches have been developed to compute integrated information \cite{arsiwalla2013iit}, \cite{arsiwalla2016computing}, \cite{arsiwalla2016high},   \cite{xda2016global},  \cite{ay2015information}, \cite{bt}, \cite{bs}, \cite{griffith2014principled},  \cite{oizumi2014}  (see also \cite{arsiwalla2009entropy},   \cite{petersen2015},    \cite{tegmark2016improved},  \cite{wennekers2005stochastic} for other related measures). Some of these were constructed for networks with discrete  states, others for continuous state variables. In this paper, we will consider stochastic network dynamics with continuous state variables because this  class of networks model many biological as well as communication systems that generate multivariate time-series signals. We want to study the precise analytic relationship between the information integrated by these networks and the couplings that parametrize their structure and dynamics. It turns out that tuning the dynamical operating point of a network near  the edge of criticality leads to a high rate of network information integration and that remains consistent across network topologies. To explain this phenomenon, we analyze  the spectrum of the network's dynamics. This  reveals that integrated information is  coupled to the characteristics of the eigenmodes of the system's covariance matrix, and in turn these are related to the eigenvalues of the network's adjacency matrix.  In this paper, we make these relationships precise.

\section{Methods}
We consider complex networks with linear multivariate dynamics and Gaussian noise. It follows that the state of each node is given by a random variable pertaining to a Gaussian distribution. For many realistic applications, Gaussian-distributed variables are fairly reasonable abstractions. The state of the network ${\bf X_t}$ at time $t$ is taken as a multivariate Gaussian variable with distribution ${\bf P_{X_t} (x_t) }$. ${\bf  x_t}$ denotes an instantiation of ${\bf X_t}$ with components ${ x_t^i }$ ($i$ going from $1$ to $n$, n being the number of nodes). When the network makes a transition from an initial state  ${\bf X_0}$ to a state ${\bf X_1}$ at time $t = 1$, observing the final state generates information about the system's initial state.  The information generated equals the reduction in uncertainty regarding the initial state ${\bf X_0}$. This is given by the conditional entropy ${ \bf H (X_0 | X_1) }$. In order to extract that part of the information generated by the system as a whole, over and above that generated individually by its irreducible parts, one computes the relative conditional entropy  given by the Kullback-Leibler divergence of the conditional distribution ${\bf P_{X_0 | X_1 = x'} (x) }$ of the whole system with respect to the joint conditional distributions $\prod_{k=1}^{n}  {\bf P_{M^k_0 | M^k_1 = m'}  }$ of its irreducible parts \cite{xda2016global}. Here ${\bf M^k_t}$ denotes the state variable of the $k^{th}$-component of the partitioned system at time $t$.  Denoting the Kullback-Leibler divergence of the above quantities as  ${ \Phi }$, we have   
\begin{eqnarray}
{ \Phi }  ( {\bf X_0 \rightarrow X_1 = x'}  )   =  \,  D_{KL} \left(   {\bf P_{X_0 | X_1 = x'}  }  \big|\big|  \prod_{k=1}^{n}  {\bf P_{M^k_0 | M^k_1 = m'}  }  \right) 
\label{KLdiv}
\end{eqnarray}
where for the partitioned system, the state variables ${\bf X_0}$ and ${\bf X_1}$ can be   expressed as formal sums  ${\bf  X_0 =  \bigoplus_{k = 1}^{n}  M_0^k  }$  and  ${\bf  X_1 =  \bigoplus_{k = 1}^{n}  M_1^k  }$ respectively. To have a measure that is independent of any particular instantiation of the final state ${\bf x'}$, we average eq.(\ref{KLdiv}) with respect to final states to obtain    
\begin{equation}
\left<   \Phi  \right>  ( {\bf X_0 \rightarrow X_1 }  )  =   -   {  \bf H  (X_0 | X_1) }  +   \sum_{k=1}^{n}  { \bf H (M^k_0 | M^k_1) }
\label{infointr}
\end{equation}
This is the definition of integrated information that we will use \cite{xda2016global}.  
%Note that  \cite{bt} do not work with networks having stochastic dynamics.  They do use eq.(\ref{KLdiv}) as their definition but endow their nodes with only binary states. On the other hand,  \cite{bs} uses a different definition of integrated information, where conditional entropies as in eq.(\ref{infointr}) are replaced by conditional mutual information. This definition only matches the definition of  eq.(\ref{KLdiv})  in special cases but not in general for any distribution. From an information theory perspective, the Kullback-Leibler divergence offers a principled way of comparing probability distributions, hence we follow that approach in formulating our measure in eq.(\ref{infointr}).  
The state variable at each time $t = 0$ and  $t = 1$ follows a multivariate Gaussian distribution   
${\bf X_0  \sim   {\cal N} \left(  \bar{x}_0, \Sigma (X_0)  \right)}$ and  ${\bf X_1  \sim   {\cal N} \left(  \bar{x}_1, \Sigma (X_1)  \right)}$ respectively.  The generative model for this system is equivalent to a multi-variate auto-regressive process      \cite{barrett2010}        
\begin{equation}
{\bf X_1 =  {\cal A} \;  X_0  + E_1 }   
\label{genmod}
\end{equation}
where ${\cal A}$ is the weighted adjacency matrix of the network and $E_1$ is Gaussian noise.  Taking the mean and  covariance respectively on both sides of this equation, while holding the residual independent of the regression variables gives 
\begin{eqnarray}
{\bf   \bar{x}_1  =    {\cal A} \;    \bar{x}_0   }  \quad  \qquad   {\bf  \Sigma(X_1)  =   {\cal A}  \;   \Sigma(X_0)  \;  {\cal A}^T  +  \Sigma(E)    }     
\label{DTLeq}   
\end{eqnarray}
In the absence of any external inputs, stationary solutions of a stochastic linear dynamical system as in eq.(\ref{genmod}) are fluctuations about the origin. Therefore, we can shift coordinates to set the means ${\bf  \bar{x}_0 }$ and consequently ${\bf  \bar{x}_1 }$ to the zero. The second equality in eq.(\ref{DTLeq}) is the discrete-time Lyapunov equation and its solution will give us the covariance matrix of the state variables. The conditional entropy for a multivariate Gaussian variable was computed in \cite{bs} 
\begin{equation}
{\bf H (X_0 | X_1)  }  =   \frac{1}{2}  n  \log (2 \pi e)  - \frac{1}{2}   \log \left[  \det    {\bf \Sigma (X_0 | X_1) }  \right]   
\end{equation}
and depends on the conditional  covariance matrix. Substituting in eq.(\ref{infointr}) yields 
\begin{equation}
\left<   \Phi  \right>  ( {\bf X_0 \rightarrow X_1 }  )   =  \frac{1}{2}   \log \left[  \frac{ \prod_{ {\bf k} = 1}^{n}  \det   {\bf \Sigma (M^k_0 | M^k_1) }  }{ \det    {\bf \Sigma (X_0 | X_1) }  }  \right]    
\label{ginfo}
\end{equation}
In order to compute the conditional covariance matrix we make use of the identity  (proof of this identity for the Gaussian case was demonstrated in \cite{barrett2010}) 
\begin{equation}
 {\bf \Sigma (X | Y)  =  \Sigma(X) -  \Sigma (X,  Y)  \Sigma (Y)^{-1}  \Sigma (X, Y)^T    }    
\label{covid}
\end{equation}
Computing ${\bf \Sigma (X_0,  X_1)  =  \Sigma (X_0)  \,  {\cal A}^T}$ and using the above identity, we get
\begin{eqnarray}
{\bf \Sigma (X_0 | X_1)}  &=& {\bf  \Sigma(X_0)  -  \Sigma (X_0)  \,  {\cal A}^T \,  \Sigma (X_1)^{-1}    {\cal A}  \;  \Sigma (X_0)^T  }    
\label{Xcondcov} \\ 
 {\bf \Sigma (M^k_0 | M^k_1) }   &=&  {\bf   \Sigma(M_0^k) }  -   {\bf  \Sigma(M_0^{k})  \,  {{\cal A}^T} \big{|}_{k}  \,  { \Sigma(M_1^{k})  }^{-1}   {\cal A}  \big{|}_{k}  \,  {\Sigma (M_0^k) }^T    }
\label{Mcondcov}
\end{eqnarray}
the conditional covariance for the whole system and that for its parts respectively. The variable ${\bf M_0^k}$ refers to the state of the $k^{th}$ node at $t = 0$ and $ {\cal A}  \big{|}_{k}$ denotes the  (trivial) restriction of the adjacency matrix to the $k^{th}$ node. For discrete-time linear systems, a unique fixed point exists as long as all eigenvalues of  ${\cal A}$ differ from one (in the absence of external inputs).  Stable stationary solutions of this  system are obtained when the magnitude of all eigenvalues is less than one. In that regime, the multivariate probability distribution of states approaches stationarity and the  covariance matrix converges, such that ${\bf \Sigma (X_1) = \Sigma (X_0)}$ (here $t = 0$ and $t = 1$ refer to time-points after the system has converged to its fixed point). Then the discrete-time Lyapunov equation in eq.(\ref{DTLeq}) can be solved for the stable covariance matrix ${\bf \Sigma (X_0)}$. For networks with symmetric adjacency matrix and independent Gaussian noise, the solution takes a particularly simple form 
\begin{eqnarray}
{\bf \Sigma (X_0)  =  \left( 1 -  {\cal A}^2 \right)^{-1}  \Sigma(E)  }      
\label{covsol}
\end{eqnarray}
and for the parts, we have
\begin{eqnarray}
{\bf  \Sigma(M_0^k)  =  \Sigma (X_0)  \big{|}_{k}    }      
\label{covparts}
\end{eqnarray}
given by the restriction of the full covariance matrix on the $k^{th}$ component. Note that eq.(\ref{covparts}) is not the same as taking eq.(\ref{covsol}) on the restricted adjacency matrix as that would mean that the $k^{th}$ node has been explicitly severed from the rest of the network. In fact,  eq.(\ref{covparts}) is the variance of the $k^{th}$ node while it is still part of the network  and $\left< \Phi \right>$ yields the amount of information that is still greater than that of the sum of these connected parts.  Inserting eqs.(\ref{Xcondcov}), (\ref{Mcondcov}), (\ref{covsol}) and (\ref{covparts}) into eq.(\ref{ginfo}) yields $\left< \Phi \right>$ as a function of network weights for symmetric  networks\footnote{For the case of asymmetric weights, the entries of the covariance matrix cannot be explicitly expressed as a matrix equation. However, they may still be solved by Jordan decomposition of both sides of the Lyapunov equation.}.

\section{Results}
Using the mathematical tools described above, we now compute exact analytic solutions for $\left< \Phi \right>$ for the 6 networks shown in Figure~\ref{fig1} below. Each of these networks have 10x10 adjacency matrices with bi-directional weights (though our analysis applies to directed graphs as well). We want to determine the  characteristics of $\left< \Phi \right>$ as a function of network weights, which we keep as free parameters. However, in order to constrain the space of parameters, we set all weights to a single parameter, the global coupling strength $g$. This coupling parameter allows us to scale network weights from zero until the point that the system's dynamics becomes unstable. This gives  us $\left< \Phi \right>$ as a function of $g$.  The analytic results for each network labeled from 1 to 6 are shown in eqs.(\ref{pnetA}), (\ref{pnetE}), (\ref{pnetD}), (\ref{pnetF}), (\ref{pnetC}) and (\ref{pnetB})  respectively.  These are computed for a time-step when the system transitions from ${\bf X_0}$ to ${\bf X_1}$. Our calculations of  $\left< \Phi \right>$ are performed for stable stationary solutions (which we can explicitly obtain from the dynamics).        
\begin{figure}[h]
\centering
\includegraphics[height=8.0cm]{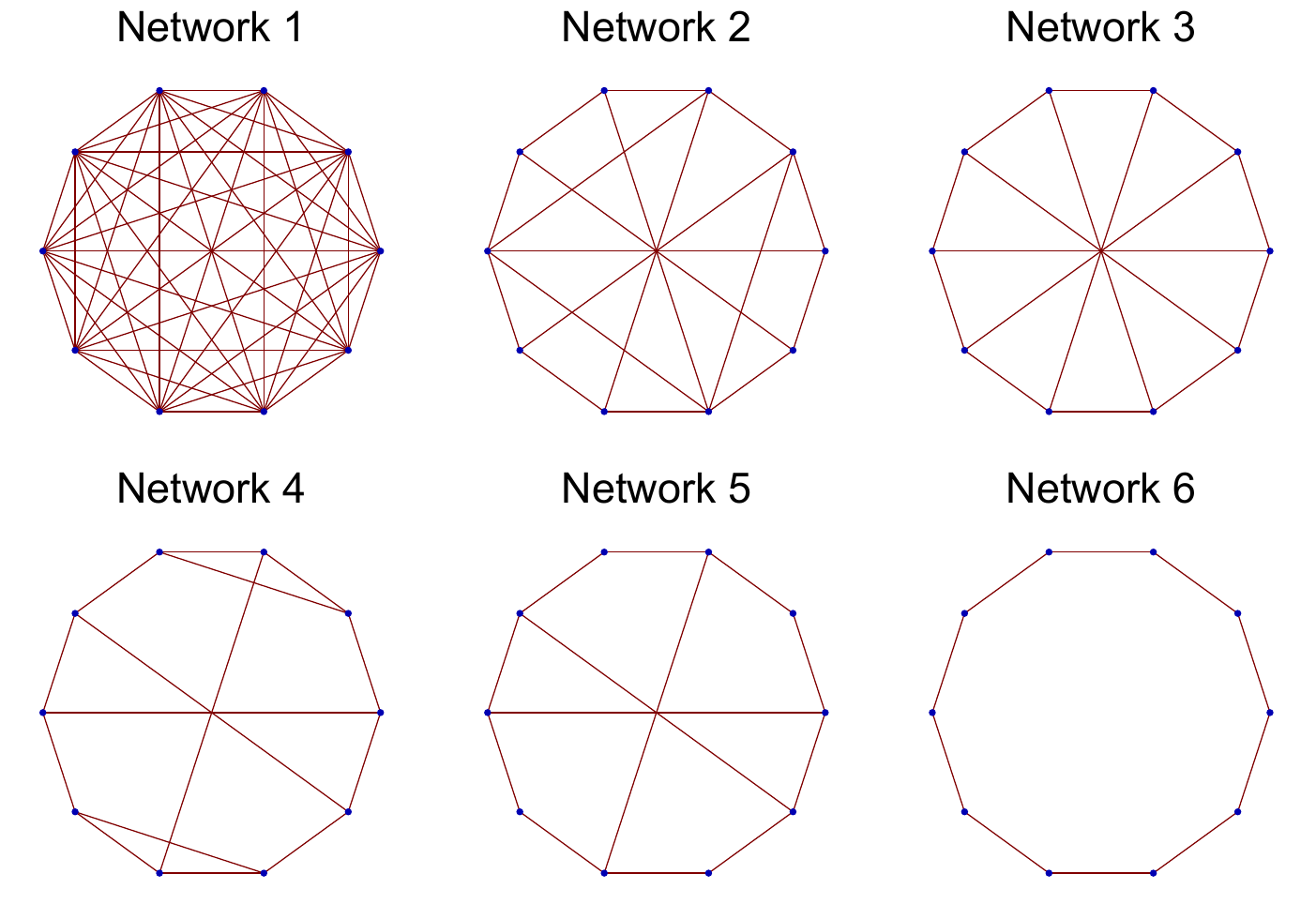}
\caption{{\bf Network topologies.} Graphs of 6 networks, from the most densely connected (Network 1) to the least densely connected (Network 6).}
\label{fig1}
\end{figure}  
\begin{eqnarray}
\left<  \Phi  \right>_1 &=&  \frac{1}{2}   \log  \frac{ \left(1 - 73 g^2 \right)^{10} }{ \left(1 - 82 g^2 + 81 g^4 \right)^{10} }     \label{pnetA}  \\ 
\left<  \Phi  \right>_2 &=&  \frac{1}{2}   \log  \frac{A_1 \cdot A_2 \cdot A_3 }{ \left(-1+18 g^2-59 g^4+59 g^6-15 g^8 + g^{10}  \right)^{10} }    \hspace{4.8cm}   
\label{pnetE}   
\end{eqnarray}
\begin{eqnarray}
\mbox{where}  \quad  A_1 &=&  \left(1-15 g^2 + 32 g^4 -19 g^6+2 g^8 \right)^2  \left(1-15 g^2+35 g^4-16 g^6+2 g^8 \right)^2   \nonumber  \\
A_2 &=&  \left(1-13 g^2+37 g^4-30 g^6+3 g^8 \right)^2  \left(1-15 g^2+37 g^4-27 g^6+4 g^8 \right)^2    \nonumber  \\
A_3 &=&  \left(1-14 g^2+36 g^4-26 g^6+4 g^8 \right)^2    \nonumber  
\end{eqnarray}  
\begin{eqnarray} 
\left<   \Phi  \right>_3 &=&  \frac{1}{2}   \log  \frac{ \left(1-9 g^2+11 g^4 \right)^{10} }{ \left(1-12 g^2+28 g^4-9 g^6 \right)^{10} }   
\label{pnetD}   \\  
\left<   \Phi  \right>_4 &=&  \frac{1}{2}   \log   \frac{B_1 \cdot B_2 \cdot B_3}{\left(1 - 3 g^2 + g^4\right)^{10}  \left(1 - 21 g^2 + 152 g^4 - 445 g^6 + 441 g^8 \right)^{10} }  \hspace{3cm}  
\label{pnetF}   
\end{eqnarray}  
\begin{eqnarray}
\mbox{where}  \quad  B_1 &=&  \left(1 - 21 g^2 + 161 g^4 - 563 g^6 + 895 g^8 - 517 g^{10} \right)^4  \nonumber  \\
B_2 &=&  \left(1 - 21 g^2 + 159 g^4 - 535 g^6 + 793 g^8 - 409 g^{10} \right)^4     \nonumber  \\
B_3 &=&  \left(1 - 21 g^2 + 159 g^4 - 533 g^6 + 769 g^8 - 355 g^{10} \right)^2      \nonumber  
\end{eqnarray}  
\begin{eqnarray}
\left<  \Phi  \right>_5 &=& \frac{1}{2}   \log  \frac{ \left(1 - 8 g^2 + 11 g^4 \right)^2  \left(1 - 10 g^2 + 26 g^4 - 19 g^6 \right)^4 \left(1 - 9  g^2 + 22 g^4 - 16 g^6 \right)^4 }{ \left(- 1 + g^2 \right)^8 \left(1 - 11 g^2 + 31 g^4 - 25 g^6\right)^{10} }   \qquad  
\label{pnetC}  \\ 
\left<   \Phi  \right>_6 &=&   \frac{1}{2}   \log  \frac{ \left(1-5 g^2+5 g^4 \right)^{10} }{ \left(1-7 g^2+13 g^4-4 g^6 \right)^{10} }   
\label{pnetB}   
\end{eqnarray}
 
\begin{figure}[h]
\centering
\includegraphics[height=6.0cm]{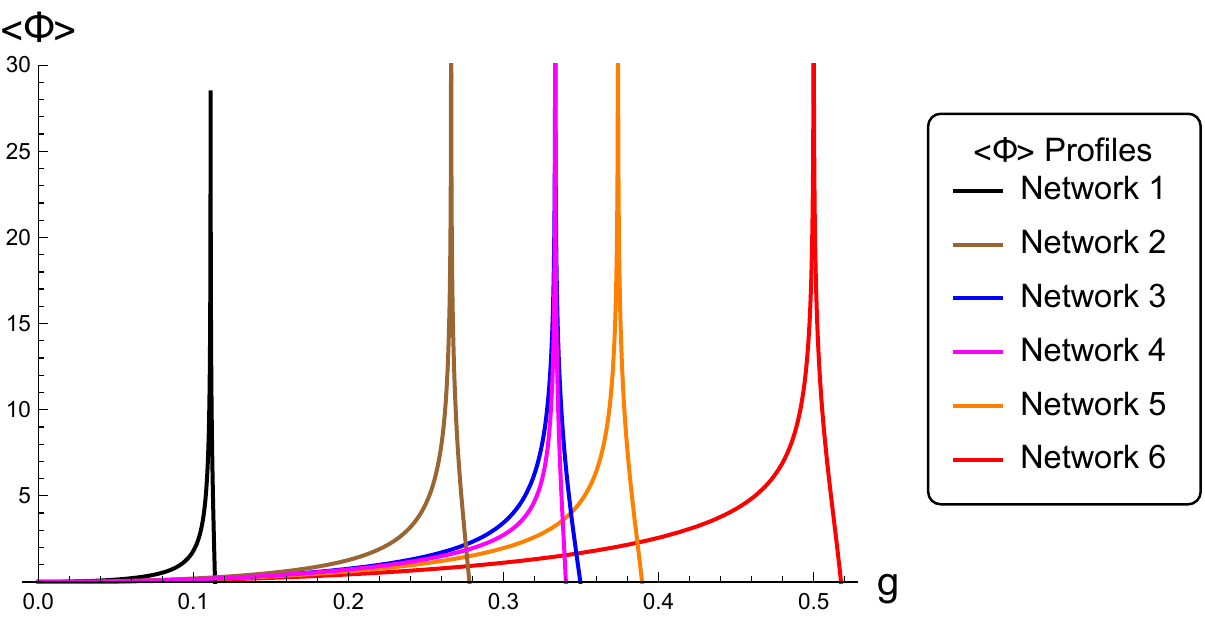}
\caption{{\bf  Analytic solutions of $\left< \Phi \right>$.}  $\left< \Phi \right>$ profiles for the networks in figure~\ref{fig1} above. These profiles display an ordering corresponding to the most densely connected network on the left to the least densely connected one on the right.}
\label{fig2}
\end{figure}

In figure~\ref{fig2} we plot characteristic $\left< \Phi \right>$ profiles for each network, based on the above solutions. This figure highlights a couple of interesting features about integrated information. First of all, irrespective of topology, all networks approach a pole at some value of $g$, near which, the integrated information of that network is extremely high. Furthermore,  the location of the pole is precisely the critical point after which the largest eigenvalue of the network slips outside of the radius of stability. However, differences in network topologies do play a role in placing each network's $\left< \Phi \right>$ profile in distinct regions of the coupling phase space. Figure~\ref{fig2} shows an ordering of these profiles: the most densely packed networks lie towards the left end, while the least densely connected ones are more on the right.  

\subsection{Spectral Modes} 
We have seen that the analytic solutions for  $\langle  \Phi  \rangle $ obtained above have poles at specific values of $g$. These poles of complexity are located at critical points of the network's dynamics.  Since the expressions we have computed are exact analytic, we can analyze  stability in the vicinity of these poles by analyzing the spectral eigenmodes of the network.  The eigenvalues of the adjacency matrix depend on the coupling $g$ (because $g$ scales the weights of the edges). Stability of the system in the stationary regime is guaranteed if and only if the norm of every eigenvalue of  ${\cal A}$ is less than one. Unit norm  corresponds to critical dynamics. Examining each of the networks considered above, we find  that these poles are located precisely at those values of the coupling $g$, where one or more eigenvalues equal unit length (shown in table \ref{table21}). From the plots of $\langle  \Phi  \rangle $ we find that irrespective of topology, network integrated information sharply increases near  criticality (while still in the stable regime).  Solutions of $\langle  \Phi  \rangle $ corresponding to stable stationary dynamics  exist for  values of coupling $g$ for which all eigenvalues of ${\cal A}$ are within the interval $(-1,1)$.  As the eigenvalues increase with increasing value of $g$, when the largest eigenvalue touches plus or minus one, the dynamical system becomes critical and hits a pole in $\langle  \Phi  \rangle $. Therefore, stable solutions are only defined for $g$ from 0 until the point where the first pole appears.  After that point, stationary solutions do not exist, though non-stationary ones may be found. In this work, we only consider    stationary solutions as these refer to fixed points/attractors of dynamical systems. Non-stationary solutions may be interesting for studying meta-stable or transient states. 

\begin{table}[htp]
	\begin{centering}
		\begin{tabular}{|c|c|c|c|c|c|}
\hline
Network 1  &   Network 2  & Network 3  & Network 4  & Network 5  & Network 6     \\
\hline  
0.111111 & 0.266223 &  0.333333 & 0.333333 &  0.373813 &  0.5  \\
1            & 0.659656 &  0.618034 & 0.404394  & 0.649693  & 0.618034 \\ 
              & 0.892236 & 1.61803    &  0.51668   & 0.823506  & 1.61803 \\
              & 2.08888   &                  & 0.618034 & 1               &   \\
              & 3.05524   &                  & 0.683715 &                 &    \\
              &                &                  &   1.61803  &                 &   \\
\hline
\end{tabular}
		\caption{Pole positions of  $\langle  \Phi  \rangle $ along the $g$ axis for all six networks.  }
		\label{table21}
	\end{centering}
\end{table}

\begin{table}[htp]
	\begin{centering}
		\begin{tabular}{|c|c|c|c|c|c|}
\hline
Network 1  &   Network 2  & Network 3  & Network 4  & Network 5  & Network 6     \\
\hline  
9$g$   &    3.75626$g$ &              3$g$ &             3$g$  &  2.67513$g$   &              2$g$   \\
-$g$    &   -3.75626$g$ &            -3$g$ & -2.47283$g$   & -2.67513$g$   &              -2$g$  \\ 
-$g$    &    1.51594$g$ &   1.61803$g$ &  1.93543$g$   &  1.53919$g$   &    1.61803$g$  \\
-$g$    &   -1.51594$g$ &   1.61803$g$ & -1.61803$g$   & -1.53919$g$  &    1.61803$g$   \\
-$g$    &    1.12078$g$ &  -1.61803$g$ & -1.61803$g$   &  1.21432$g$  &   -1.61803$g$  \\
-$g$    &   -1.12078$g$ &  -1.61803$g$ &  1.61803$g$   & -1.21432$g$  &   -1.61803$g$  \\
-$g$    &  0.478725$g$ &  0.618034$g$ &  -1.4626$g$   &               $g$   &  0.618034$g$  \\
-$g$    & -0.478725$g$ &  0.618034$g$ & -0.618034$g$  &               $g$  &  0.618034$g$  \\
-$g$    &  0.327307$g$ & -0.618034$g$ & 0.618034$g$   &              -$g$  & -0.618034$g$   \\
-$g$    & -0.327307$g$ & -0.618034$g$ & 0.618034$g$   &              -$g$  & -0.618034$g$  \\
\hline 
\end{tabular}
		\caption{Network spectra showing eigenvalues of ${\cal A}$ for all six networks.   }
		\label{table22}
	\end{centering}
\end{table}

\begin{figure}[h]
\begin{center}
\begin{minipage}[h]{0.5\linewidth} 
\centering
\includegraphics[width=0.95\linewidth]{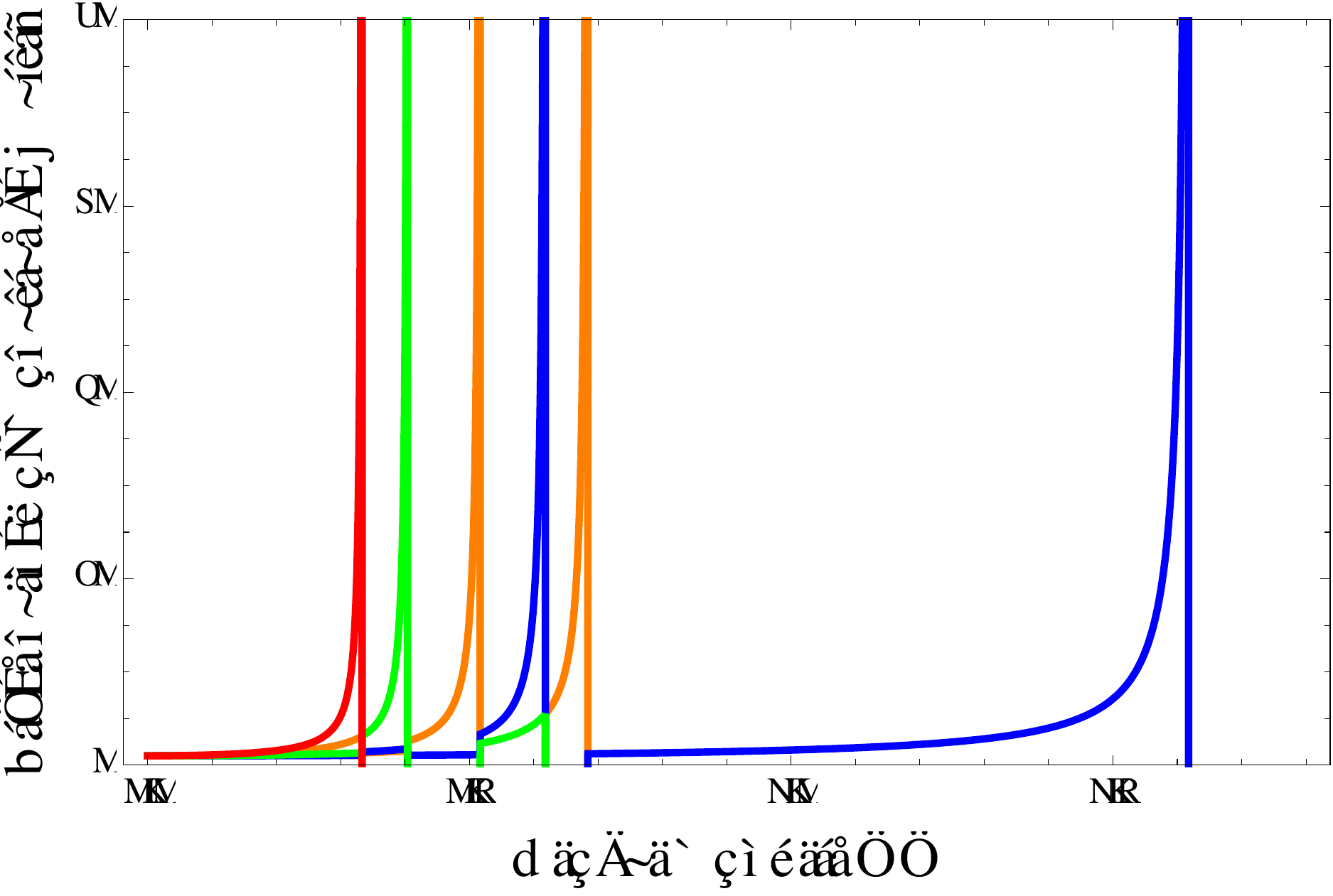} 
\end{minipage}%
\begin{minipage}[h]{0.5\linewidth} 
\centering
\includegraphics[width=0.95\linewidth]{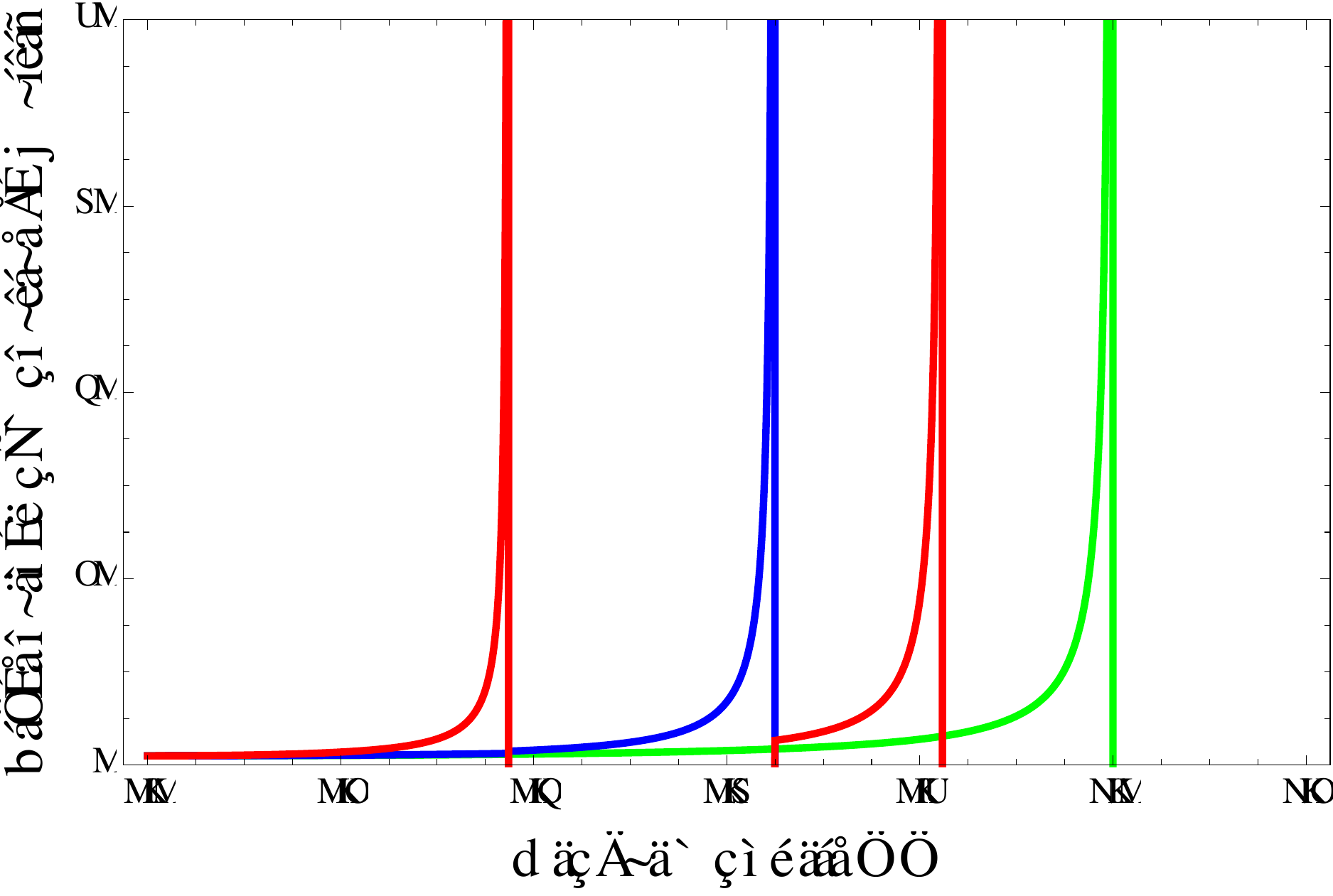} 
\end{minipage}
\end{center}
\caption{ {\bf Eigenvalues of the covariance matrix ${\bf \Sigma(X)}$. } Eigenvalue profiles shown as a function of coupling $g$ for network 4 (left) and network 5 (right). Each color denotes an eigenvalue. Some eigenvalues are degenerate and some have multiple poles. The eigenvectors of ${\bf \Sigma(X)}$ are the principle modes of the time-series ${\bf X}$ and the corresponding eigenvalues shown above indicate the variance of these eigenmodes. A large variance implies a dominance of that mode. We see that dynamics of the system near the critical coupling $g$ for each system is dominated by the leading eigenmode (shown in red). }
\label{figeig}
\end{figure}

From the computed expressions  of  $\langle  \Phi  \rangle$, we determine the position of the poles by computing the roots of the denominators. For our six networks, these poles occur at the values of $g$ shown in table \ref{table21}. Now consider the eigenmodes of the adjacency matrix ${\cal A}$. As each edge has weight $g$, the eigenvalues of ${\cal A}$ are multiples of the coupling $g$ and are shown for all networks in table \ref{table22}.  Having this, if we now insert the value of $g$ at the first pole for each network into its corresponding eigenvalues, we find that the largest eigenvalue becomes precisely plus or minus 1 at the first pole and subsequent eigenvalues have norms less than one. After the first pole is crossed by increasing $g$, the norm of the largest eigenvalue crosses unit length and subsequent other eigenvalues start growing as well. The second pole is reached, precisely when the second largest eigenvalue hits norm 1 and so on until beyond the last pole when all the eigenvalues have length greater than 1.   To understand what happens  to the dynamics when  eigenvalues of ${\cal A}$ approach 1, we examine eq.(\ref{covsol}). This tells us that unit eigenvalues of ${\cal A}$ lead to divergences in the covariance matrix, thus  leading to  poles in the corresponding eigenvalues of the covariance matrix. This is shown in figure~\ref{figeig} below for network 4 and network 5.     The eigenvectors of the covariance matrix are the principle modes of the time-series ${\bf X}$ and the corresponding eigenvalues give the variance of these modes. A large variance implies a dominance of that mode. Therefore, when the system approaches a critical point, the dynamics is completely dominated by that specific mode. The  poles of $\langle \Phi  \rangle$ exactly coincide with the poles of the system's  covariance eigenmodes and as these leading modes gradually grow towards the critical point, so does $\langle \Phi \rangle$.    Also, because stable stationary solutions only exist for $g$ until the first critical point, we see that after the first critical point, positivity of eigenvalues of the covariance matrix is not guaranteed. We have checked this numerically.  This helps identify the full range of stable  solutions of $\langle \Phi  \rangle$  shown in figure~\ref{fig2}.       

% Thus, the integrated information of stochastic dynamic networks is coupled to the characteristics of the eigenmodes of its covariance matrix, which in turn are related to the eigenvalues of the network's connectivity since the location of all  poles coincide with critical points of the connectivity matrix ${\cal A}$. 

\section{Discussion}
This paper analytically investigates why informational complexity of complex networks with stochastic Gaussian dynamics shows a characteristic strong growth near  criticality, and one which persists across different network topologies. We have used integrated information or $\left< \Phi \right>$ as a candidate complexity measure as its formulation elegantly makes use of both, the network's structure and dynamics, while serving as a global measure for the system's collective states.  We have computed  exact analytic solutions for $\left< \Phi \right>$ as a function of the network's overall coupling parameter $g$.   Concurrent to computing $\left< \Phi \right>$, we have also analyzed the spectral properties of the network, which shows how the network's collective eigenmodes contribute to $\left< \Phi \right>$.      We found poles in  $\left< \Phi \right>$ at point of criticality,  leading to high information integration near the edge of criticality. This indicates that it is not only the network's topology that determines how much information it can integrate, but also its dynamical operating point. As a matter of fact, operating near the edge of criticality (when still within the stable regime)  leads to a sharp increase in $\left< \Phi \right>$, irrespective of network topology. 

Looking at the eigenmodes of the network, we found that instances of high integrated information (in the stationary regime) are associated to the strong dominance of the eigenmode corresponding to the leading eigenvalue of the network's covariance matrix, while sub-leading modes get suppressed. High complexity corresponds  to a symphony-like state of the network, rather than a globally synchronized state.  Hence,  $\left< \Phi \right>$  taken as a proxy for a  system's information processing capacity,   implies that operating at the edge of criticality can be favorable for efficient information processing.  
This can be particularly beneficial in designing communication networks and also in understanding information processes in biological networks such as the human brain. And in fact, there has been recent evidence precisely in this direction \cite{deco2013resting}.  The authors of \cite{deco2013resting}  demonstrate that  resting-state dynamics of the human brain network in fact operates precisely at the edge of a bifurcation (this result also holds for higher resolution datasets   \cite{arsiwalla2015network},  \cite{arsiwalla2015connectomics}).  Combining this observation with the results of this   paper suggests that the reason why  the dynamics of the brain might operate at the edge of criticality is to enable efficient information integration.    

\subsection{Acknowledgments}
\label{sect:acks}
This work has been supported by the European Research Council's CDAC project: "The Role of Consciousness in Adaptive Behavior: A Combined Empirical, Computational and Robot based Approach" (ERC-2013- ADG 341196).   
% EC FP7  grant agreement n. 258749 
%- The Role of Consciousness in Adaptive Behavior: A Combined Empirical, Computational and Robot based Approach (ERC-2013- ADG 341196).  

%------------------------------------------------------------------------------
% Refs:
%
\label{sect:bib}
\bibliographystyle{plain}
\bibliography{test2}

\begin{thebibliography}{10}

\bibitem{arsiwalla2013iit}
X.~D. Arsiwalla and P.~F. M.~J. Verschure.
\newblock Integrated information for large complex networks.
\newblock In {\em The 2013 International Joint Conference on Neural Networks
  (IJCNN)}, pages 1--7, Aug 2013.

\bibitem{arsiwalla2009entropy}
Xerxes~D Arsiwalla.
\newblock Entropy functions with 5d chern-simons terms.
\newblock {\em Journal of High Energy Physics}, 2009(09):059, 2009.

\bibitem{arsiwalla2015connectomics}
Xerxes~D Arsiwalla, David Dalmazzo, Riccardo Zucca, Alberto Betella, Santiago
  Brandi, Enrique Martinez, Pedro Omedas, and Paul Verschure.
\newblock Connectomics to semantomics: Addressing the brain's big data
  challenge.
\newblock {\em Procedia Computer Science}, 53:48--55, 2015.

\bibitem{arsiwalla2016computing}
Xerxes~D. Arsiwalla and Paul Verschure.
\newblock {\em Computing Information Integration in Brain Networks}, pages
  136--146.
\newblock Springer International Publishing, Cham, Switzerland, 2016.

\bibitem{arsiwalla2016high}
Xerxes~D. Arsiwalla and Paul F. M.~J. Verschure.
\newblock {\em High Integrated Information in Complex Networks Near
  Criticality}, pages 184--191.
\newblock Springer International Publishing, Cham, Switzerland, 2016.

\bibitem{xda2016global}
Xerxes~D Arsiwalla and Paul~FMJ Verschure.
\newblock The global dynamical complexity of the human brain network.
\newblock {\em Applied Network Science}, 1(1):16, 2016.

\bibitem{arsiwalla2015network}
Xerxes~D. Arsiwalla, Riccardo Zucca, Alberto Betella, Enrique Martinez, David
  Dalmazzo, Pedro Omedas, Gustavo Deco, and Paul Verschure.
\newblock Network dynamics with brainx3: A large-scale simulation of the human
  brain network with real-time interaction.
\newblock {\em Frontiers in Neuroinformatics}, 9(2), 2015.

\bibitem{ay2015information}
Nihat Ay.
\newblock Information geometry on complexity and stochastic interaction.
\newblock {\em Entropy}, 17(4):2432--2458, 2015.

\bibitem{bt}
David Balduzzi and Giulio Tononi.
\newblock Integrated information in discrete dynamical systems: motivation and
  theoretical framework.
\newblock {\em PLoS Comput Biol}, 4(6):e1000091, 2008.

\bibitem{barrett2010}
Adam~B Barrett, Lionel Barnett, and Anil~K Seth.
\newblock Multivariate granger causality and generalized variance.
\newblock {\em Physical Review E}, 81(4):041907, 2010.

\bibitem{bs}
Adam~B Barrett and Anil~K Seth.
\newblock Practical measures of integrated information for time-series data.
\newblock {\em PLoS Comput Biol}, 7(1):e1001052, 2011.

\bibitem{deco2013resting}
Gustavo Deco, Adri{\'a}n Ponce-Alvarez, Dante Mantini, Gian~Luca Romani, Patric
  Hagmann, and Maurizio Corbetta.
\newblock Resting-state functional connectivity emerges from structurally and
  dynamically shaped slow linear fluctuations.
\newblock {\em The Journal of Neuroscience}, 33(27):11239--11252, 2013.

\bibitem{griffith2014principled}
Virgil Griffith.
\newblock A principled infotheoretic$\backslash$ phi-like measure.
\newblock {\em arXiv preprint arXiv:1401.0978}, 2014.

\bibitem{oizumi2014}
Masafumi Oizumi, Larissa Albantakis, and Giulio Tononi.
\newblock From the phenomenology to the mechanisms of consciousness: integrated
  information theory 3.0.
\newblock {\em PLoS Comput Biol}, 10(5):e1003588, 2014.

\bibitem{petersen2015}
Karl Petersen and Benjamin Wilson.
\newblock Dynamical intricacy and average sample complexity.
\newblock {\em arXiv preprint arXiv:1512.01143}, 2015.

\bibitem{tegmark2016improved}
Max Tegmark.
\newblock Improved measures of integrated information.
\newblock {\em arXiv preprint arXiv:1601.02626}, 2016.

\bibitem{tononi2004}
Giulio Tononi.
\newblock An information integration theory of consciousness.
\newblock {\em BMC neuroscience}, 5(1):42, 2004.

\bibitem{tononi2003}
Giulio Tononi and Olaf Sporns.
\newblock Measuring information integration.
\newblock {\em BMC neuroscience}, 4(1):31, 2003.

\bibitem{tononi1994}
Giulio Tononi, Olaf Sporns, and Gerald~M Edelman.
\newblock A measure for brain complexity: relating functional segregation and
  integration in the nervous system.
\newblock {\em Proceedings of the National Academy of Sciences},
  91(11):5033--5037, 1994.

\bibitem{wennekers2005stochastic}
Thomas Wennekers and Nihat Ay.
\newblock Stochastic interaction in associative nets.
\newblock {\em Neurocomputing}, 65:387--392, 2005.

\end{thebibliography}

%------------------------------------------------------------------------------
% Index
%\printindex

%------------------------------------------------------------------------------
\end{document}